\documentclass{article}

\usepackage{booktabs}
\usepackage{caption}
\usepackage{multirow}
\usepackage{amssymb}
\usepackage{graphicx}
\newcommand{\cmark}{\ensuremath{\checkmark}}
\usepackage[final]{corl_2026} % Uncomment for the camera-ready ``final'' version.

\hypersetup{
  pdftitle={From Foundation VLM to Versatile Driving VLM: A Triplet Incremental Training Approach},
  pdfauthor={Shengtao Wu, Zichao Dong, Weixuan Qiao, Xiaolu Liu, Xin Zhan, Junbo Chen, Jianke Zhu}
}

\title{From Foundation VLM to Versatile Driving VLM:\\ A Triplet Incremental Training Approach}

\author{
  \textbf{Shengtao Wu}\textsuperscript{\mdseries 1*}, \textbf{Zichao Dong}\textsuperscript{\mdseries 2*\textdagger}, \textbf{Weixuan Qiao}\textsuperscript{\mdseries 1}, \textbf{Xiaolu Liu}\textsuperscript{\mdseries 1}, \\
  \textbf{Xin Zhan}\textsuperscript{\mdseries 2}, \textbf{Junbo Chen}\textsuperscript{\mdseries 2}, \textbf{Jianke Zhu}\textsuperscript{\mdseries 1\textdaggerdbl} \\[0.3em]
  \textnormal{\textsuperscript{\mdseries 1}Zhejiang University \qquad \textsuperscript{\mdseries 2}Udeer.AI}
}

\begin{document}
\maketitle

\begingroup
  \renewcommand{\thefootnote}{\fnsymbol{footnote}}
  \footnotetext[1]{Equal contribution.\quad\textsuperscript{\textdagger}Project Leader.\quad\textsuperscript{\textdaggerdbl}Corresponding author.}
\endgroup

\vspace{-2em}
\begin{figure}[htbp]
	\centering
	\includegraphics[width=\linewidth]{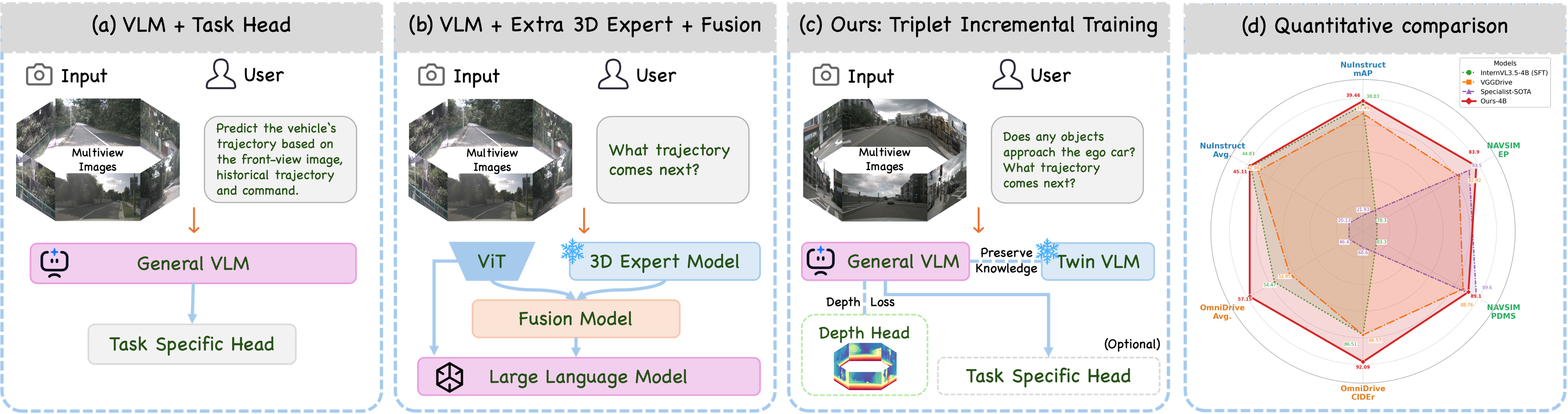}
	\caption{
		Comparison of driving VLM paradigms.
		(a) Task-head tuning paradigms suffer from fragile spatial boundaries without explicit depth supervision. 
		(b) External expert fusion approaches introduce heavy computational and inference overhead. 
		(c) Our Triplet Incremental Training aligns geometric features via a deployment-discarded transient depth branch, while regularizing semantics with a twin VLM.
		(d) Quantitative comparison demonstrating our state-of-the-art performance.
	}
	\label{fig:teaser}
\end{figure}
\vspace{-1em}

%===============================================================================
\begin{abstract}
	Adapting general Vision-Language Models (VLMs) to Autonomous Driving (AD) is thought to require heavy 3D spatial priors or intrusive architectural overhauls to compensate for a lack of spatial grounding.
    In this work, we present a counterintuitive pilot study: off-the-shelf general VLMs already exhibit remarkable adaptability across diverse driving tasks, while precise metric depth remains the primary missing link, a bottleneck readily addressable via onboard sensors and depth foundation models.
	This insight reveals that instead of complex over-engineering, general VLMs merely require a targeted geometric catalyst to bridge the spatial gap.
	To this end, we introduce the \textit{Triplet Incremental Training Framework}, a non-intrusive paradigm that seamlessly instills 3D geometric awareness into a general VLM backbone while fully preserving its open-world capabilities.
    Driven by automatically constructed (image, depth, QA pair) triplets, our framework operates via a unified \textit{tri-stream training pipeline}: an unfrozen primary VLM stream to adapt to driving tasks, an auxiliary depth stream that guides the visual backbone to implicitly imagine 3D scene geometry via explicit geometric constraints, and a frozen twin VLM stream employing logit-level KL distillation to prevent semantic drift.
	Crucially, the auxiliary depth branch is discarded during deployment, ensuring the enhanced geometric awareness comes with zero inference overhead.
	We evaluate our framework across four major benchmarks targeting four diverse capabilities spanning 3D spatial perception, driving-oriented language reasoning, end-to-end planning, and general multimodal understanding.
	Our approach outperforms state-of-the-art driving VLMs, rivals domain-specific expert models in 3D perception, and successfully retains its original general reasoning capabilities. 
	This demonstrates that Triplet Incremental Training offers a unified, deployment-ready paradigm for transforming general foundation models into robust driving intelligences.
\end{abstract}
%===============================================================================

\section{Introduction}
\label{sec:introduction}
	General Vision-Language Models (VLMs)~\citep{wang2025internvl3, zhu2025internvl3, bai2025qwen3, liu2023visual, hurst2024gpt} have emerged as powerful backbones for autonomous driving (AD)~\citep{Yang2023LLM4DriveAS, tian2024drivevlm, xu2024drivegpt4}, offering strong open-world semantic reasoning to overcome the brittleness of traditional task-specific pipelines~\citep{chitta2022transfuser, weng2024drive}. 
    Our pilot study demonstrates that off-the-shelf general VLMs already deliver highly competitive driving capabilities, with dense depth perception and 3D geometric grounding remaining as the final pieces required for safe physical interaction within driving scenes~\citep{cheng2024spatialrgpt, chen2024spatialvlm}.

	Current methodologies typically address this spatial frontier from two sub-optimal extremes. 
    Text-driven alignment approaches~\citep{li2025recogdrive, jia2023driveadapter, wang2025omnidrive} attempt to implicitly encode 3D spatial relations into text tokens via large-scale coordinate-based question-answering. 
    However, without explicit geometric constraints, these models often struggle with fine-grained critical spatial boundaries.
	Conversely, structural approaches resort to intrusive modifications, integrating heavy 3D expert branches~\citep{wang2026vggdrive} or multimodal inputs. 
	These introduce redundant computational overhead, disrupt end-to-end optimization, and risk degrading the VLM’s general capabilities through domain overfitting.

	In this work, we present an elegant alternative: \textit{general VLMs do not require complex structural overhauls to drive safely; rather, their already powerful visual representations simply require a targeted geometric catalyst.}
    We identify dense metric depth as the ideal bridge to complete this spatial consciousness. 
	Although dense absolute depth is absent from general VLM pretraining, it is readily available in AD environments---either from onboard sensors~\citep{caesar2020nuscenes} or specialized depth foundation models~\citep{hu2024metric3d, piccinelli2024unidepth, piccinelli2025unidepthv2, yang2024depth, lin2025prompting}.
	Compared with sparse text-space supervision, depth offers dense, scalable geometric signals tightly coupled with the spatial structure of the driving scene.

	Based on this insight, we introduce the \textit{Triplet Incremental Training Framework}, an incremental training paradigm built on structured (image, depth, QA pair) supervision. 
	During training, the framework optimizes the primary network via three complementary objectives.
	First, the primary VLM undergoes standard supervised fine-tuning directly on downstream driving tasks to master domain-specific capabilities.
	Concurrently, on the perception side, an auxiliary depth stream employs a lightweight depth head~\citep{ranftl2021vision} to enforce explicit geometric constraints, guiding the primary visual backbone to implicitly model 3D scene geometry.
	Meanwhile, on the semantic side, a frozen twin VLM acts as a distribution anchor, regularizing the primary VLM via off-policy logit-level KL distillation to mitigate the catastrophic forgetting of general-purpose capabilities.
	Crucially, the auxiliary depth branch is entirely discarded after training, ensuring this enhanced 3D awareness is instilled into the baseline representation with no additional architectural or inference overhead.

	To summarize, our main contributions are:
	\begin{itemize}
		\item We introduce the Triplet Incremental Training framework, a simple yet effective paradigm driven by automated (image, depth, QA pair) structures, which co-aligns explicit geometric constraints with off-policy language distillation to inject 3D spatial consciousness.
		\item We present a non-intrusive network architecture featuring a transient depth branch and a frozen twin regularizer, which completes the visual backbone's geometric alignment during training but introduces no additional architectural or inference overhead upon deployment.
		\item We conduct extensive evaluations across four distinct driving and reasoning paradigms, demonstrating state-of-the-art performance in 3D perception, language reasoning, and non-reactive trajectory planning while effectively mitigating catastrophic forgetting.
	\end{itemize}
%===============================================================================

\section{Related Work}
\label{sec:related}
\subsection{Vision-Language Models for Autonomous Driving}
	Recent autonomous-driving VLMs explore how multimodal foundation models can support scene understanding, reasoning, and planning~\citep{xu2024drivegpt4, sima2024drivelm, shao2024lmdrive, hwang2024emma, huang2025robotron, ma2024dolphins, zhao2025sce2drivex, zhang2024wisead}.
	Early systems such as DriveGPT4~\citep{xu2024drivegpt4} and DriveMLM~\citep{wang2023drivemlm} fine-tune multimodal models on driving-specific instruction data or behavioral planning states, enabling interpretable question answering and closed-loop decision making. 
	More recent resources, such as OmniDrive~\citep{wang2025omnidrive}, broaden the setting with holistic driving QA and counterfactual reasoning, while ReCogDrive~\citep{li2025recogdrive} further couples VLM reasoning with planning-oriented optimization. 
	Collectively, these works establish VLMs as a promising interface for autonomous driving, but their supervision remains largely language-centric, with limited explicit geometric grounding.

\subsection{Geometry-Aware Adaptation and Capability Preservation}
	To inject explicit 3D geometric awareness, a conventional line of work introduces heavy spatial priors or structural modifications, building on early 3D perception networks~\citep{liu2022petr, wang2022detr3d, wang2021fcos3d, wang2022probabilistic, liu2023bevfusion, li2024bevformer}. 
	For instance, VGGDrive~\citep{wang2026vggdrive} integrates an external 3D foundation expert branch via cross-view geometric enablers. 
	However, such intrusive over-engineering disrupts the structural elegance of end-to-end optimization, yields suboptimal guidance in complex scenes, and introduces massive computational overhead at inference time. 
	In parallel, recent depth foundation models~\citep{hu2024metric3d, piccinelli2024unidepth, piccinelli2025unidepthv2, yang2024depth, lin2025prompting} suggest a significantly lighter alternative, demonstrating that sparse metric prompts can guide globally consistent dense depth generation as an effective geometric catalyst~\citep{lin2025prompting}.

	Beyond geometric adaptation, preserving the native, open-world capabilities of foundation models remains a critical challenge, as standard fine-tuning is highly prone to catastrophic forgetting~\citep{sima2024drivelm, li2025recogdrive, li2017learning}. 
	While current approaches attempt to alleviate this via privileged teachers~\citep{xu2024vlm}, truncated diffusion policies~\citep{liao2025diffusiondrive}, or reinforcement learning~\citep{jiang2025alphadrive}, they often complicate deployment. 
	Different from these methods, our Triplet Incremental Training framework seamlessly injects 3D geometric awareness via an auxiliary depth stream that is entirely discarded at deployment, while a frozen twin VLM regularizes the language distribution through off-policy knowledge distillation. 
	This design effectively prevents semantic drift, transforming general foundation models into robust driving intelligences with zero inference overhead.
%===============================================================================

\section{Method}
\label{sec:method}
This section details the Triplet Incremental Training framework, a non-intrusive paradigm designed to instill robust 3D geometric awareness into VLMs while preserving the model's native open-world capabilities. 
Instead of relying on structural overhauls or text-only coordinate alignment, our approach introduces a targeted geometric catalyst during training that is completely discarded at inference. 
The framework is built upon three pillars: automated cross-modal triplet construction, a tri-stream training architecture, and a tri-objective joint optimization scheme.

\subsection{Triplet Definition and Construction}
\label{sec:triplet_definition_and_construction}
	To bridge the spatial gap without manual annotation, we reformulate downstream driving adaptation into an incremental training paradigm driven by automatically synthesized training triplets:
	\begin{equation}
	    \mathcal{T} = \left\{ \left( I, D, P \right) \right\}
	\end{equation}
	where $I \in \mathbb{R}^{H \times W \times 3}$ denotes the driving image, $D \in \mathbb{R}^{H \times W}$ is the corresponding dense metric depth map, and $P = (Q_{txt}, A_{txt})$ represents a context-aware question-answering (QA) pair.
	The automated data synthesis pipeline operates as follows:

	\textbf{Dense Metric Depth Generation.} 
	To provide fine-grained spatial boundaries, we harness the powerful zero-shot estimation capability of Prompt Depth Anything \citep{lin2025prompting}. 
	Raw sparse LiDAR point clouds are directly projected onto the image plane to serve as a sparse metric prompt $D_{sparse}$. 
	Conditioned on $D_{sparse}$, the foundation model propagates the absolute scale across the dense pixel grid in a single forward pass, efficiently producing a pixel-aligned metric depth map $D \in \mathbb{R}^{H \times W}$.

	\textbf{Spatial QA Pair Generation.} 
	Conventional driving tuning often restricts VLMs to rigid text patterns (e.g., coordinate bounding boxes or repetitive driving jargon), which warps their inherent language manifold and causes semantic collapse. 
	To circumvent this, we leverage the original foundation VLM~\citep{wang2025internvl3} to generate descriptive conversations $P = (Q_{txt}, A_{txt})$ regarding general driving scenes directly from image $I$. 
	By ensuring these QA pairs inherently conform to the foundation VLM's own expressive language distribution, they provide a pristine, self-aligned distribution anchor for the logit-level KL distillation. 
	Consequently, the primary VLM successfully assimilates 3D geometric awareness while effectively preserving its native multimodal intelligence.

\begin{figure}[t]
	\centering
	\includegraphics[width=\linewidth]{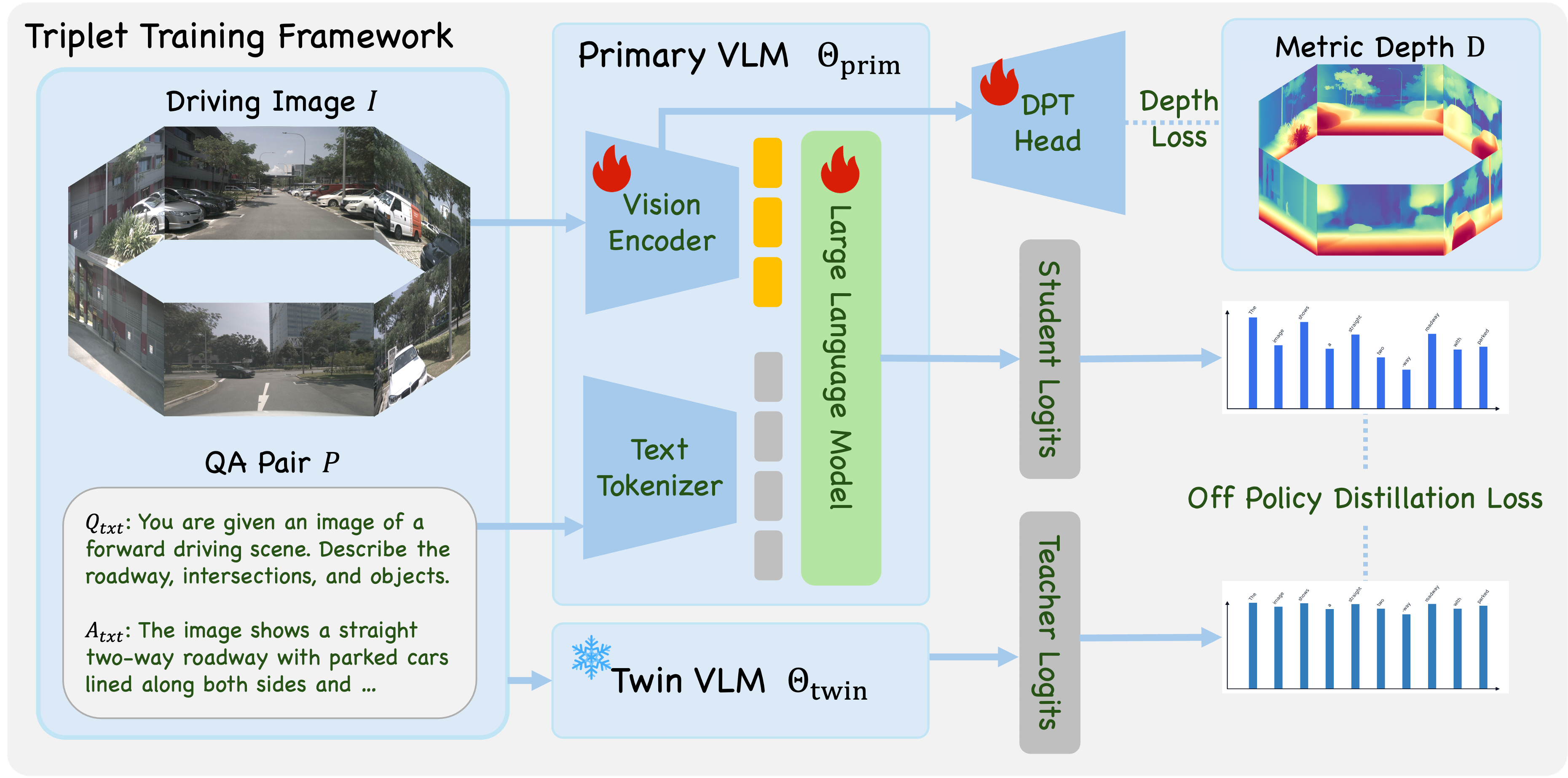}
	\caption{Overview of the Triplet Incremental Training pipeline. Specifically, the primary trainable VLM is optimized via dual constraints: a transient, deployment-discarded auxiliary depth branch for explicit geometric feature alignment, and a frozen twin VLM branch employing off-policy logit-level KL distillation to prevent semantic drift.}
	\label{fig:pipeline}
	\vspace{-1.7em}
\end{figure}

\subsection{Tri-Stream Architecture}
	To decouple spatial localization from open-world capability preservation, the framework routes the input data through three concurrent operational streams:

	\textbf{Primary VLM Stream $\Theta_{\mathrm{prim}}$:}
	This stream constitutes the core trainable network, comprising an unfrozen visual encoder and a large language model (LLM). 
	It directly absorbs task-specific language gradients and dense geometric signals to adapt to diverse autonomous driving scenarios.

	\textbf{Frozen Twin Stream $\Theta_{\mathrm{twin}}$:} 
	This stream acts as a frozen counterpart of the pretrained foundation VLM, remaining static throughout training.
	It acts as a permanent semantic anchor, regularizing the vocabulary distribution of the primary stream to prevent domain-specific text overfitting.

	\textbf{Auxiliary Depth Stream:} 
	This stream attaches a lightweight, DPT-style depth head directly to the visual encoder of $\Theta_{\mathrm{prim}}$ rather than the LLM side.
	This design aligns with the slow-fast dual-system paradigm in autonomous driving~\citep{tian2024drivevlm}, where real-time reactive perception operates independently of deliberative linguistic reasoning.
	Bypassing the heavy autoregressive language decoder ensures that dense geometric alignment directly carves the low-latency spatial representations required for reactive driving tasks.
	It projects, fuses, and upsamples multi-scale intermediate feature maps $\{f_{l}\}_{l=1}^{4}$ to yield continuous dense metric depth maps $\hat{D} = DPT(\{f_{l}\}_{l=1}^{4}) \in \mathbb{R}^{H \times W}$, acting as a transient geometric catalyst that is completely discarded during deployment.

\begin{figure}[t]
	\centering
	\includegraphics[width=0.83\linewidth]{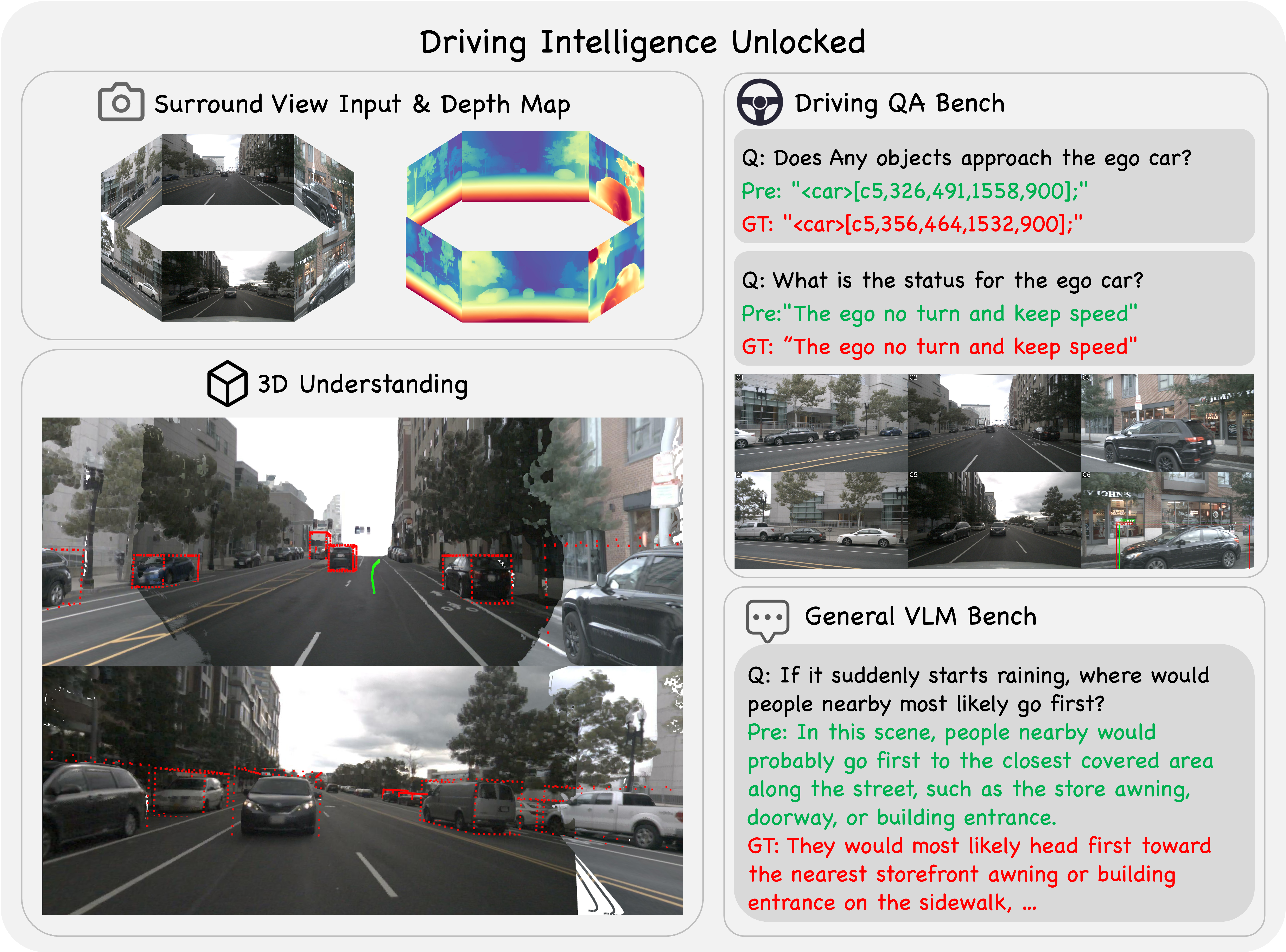}
	\caption{Visual demonstration of our model across multiple modalities and benchmarks.}
	\label{fig:experiment}
	\vspace{-1.5em}
\end{figure}

\subsection{Tri-Objective Joint Optimization}
	The primary challenge of tuning general VLMs for autonomous driving tasks requires a delicate balance between aggressive geometric adaptation and strict prevention of semantic degradation. 
	We resolve this trade-off via a tri-objective optimization scheme consisting of task adaptation, explicit geometric grounding, and distribution regularization.

	\textbf{Off-Policy Logit-Level Knowledge Distillation.} 
	To retain native intelligence while adapting to driving semantics, the language output space of the primary VLM is constrained by its twin counterpart.
	We implement an off-policy distillation scheme directly utilizing the pre-constructed training triplets.
	Specifically, given a prompt $Q_{txt}$ and image $I$, let the target answer sequence from $\mathcal{T}$ be $A_{txt} = \{t_1, t_2, \ldots, t_N\}$. 
	We compute the token-level Kullback-Leibler (KL) divergence~\citep{hinton2015distilling} strictly over the vocabulary distributions at each target token position:
	\begin{equation}
		\mathcal{L}_{\mathrm{KD}} = \frac{1}{N}\sum_{i=1}^{N}
		D_{\mathrm{KL}}\left(
		P_{\mathrm{twin}}(\cdot \mid I, Q_{\mathrm{txt}}, t_{<i}) \, \| \,
		P_{\mathrm{prim}}(\cdot \mid I, Q_{\mathrm{txt}}, t_{<i})
		\right)
	\end{equation}
	where $P(\cdot)$ represents the softmax probability distribution over the vocabulary space.
	Crucially, the prompt tokens $Q_{txt}$ are masked out, ensuring that the distillation gradient strictly anchors the generated response sequence to the foundation model's original language manifold.

	\textbf{Supervised Fine-Tuning Task Loss.}
	Concurrently, to ensure the model adapts to downstream autonomous driving tasks, we employ a standard autoregressive Cross-Entropy (CE) task loss on the target tokens of the downstream driving answers $\{y_1, y_2, \ldots, y_M\}$.
	Given a task prompt $Q_{\mathrm{task}}$:
	\begin{equation}
	    \mathcal{L}_{\mathrm{CE}} = -\frac{1}{M} \sum_{i=1}^{M} \log P_{\mathrm{prim}} \left( y_i \mid I, Q_{\mathrm{task}}, y_{<i} \right)
	\end{equation}

	\textbf{Geometric Feature Alignment.} 
	To guarantee that 3D geometric awareness is deeply embedded within the visual representations, the primary visual backbone's intermediate representations are explicitly constrained via the Scale-Invariant Logarithmic (SiLog) loss~\citep{eigen2014depth} between the predicted depth $\hat D$ and the metric depth target $D$:
	\begin{equation}
		\mathcal{L}_{\mathrm{depth}} = \sqrt{\frac{1}{|\Omega|} \sum_{p \in \Omega} e_p^2 - \lambda \left(\frac{1}{|\Omega|} \sum_{p \in \Omega} e_p\right)^2},
		\qquad e_p = \log \hat{D}_p - \log D_p .
	\end{equation}
	Here, $\Omega$ denotes the set of valid depth pixels, $e_p$ is the logarithmic depth error, and $\lambda$ controls the correction for global log-scale bias.
	The loss encourages relative geometric consistency while retaining sensitivity to absolute scale when $\lambda<1$.

	\textbf{Total Objective} 
	The complete optimization objective for the primary VLM and the attached depth head is formulated as a linear combination of the three distinct constraints:
	\begin{equation}
		\label{eq:total_objective}
		\mathcal{L} = \alpha \mathcal{L}_{\mathrm{depth}} + \beta \mathcal{L}_{\mathrm{KD}} + \gamma \mathcal{L}_{\mathrm{CE}}
	\end{equation}
	where $\alpha$, $\beta$ and $\gamma$ control the trade-off between geometric adaptation, foundation capability preservation, and task-specific fine-tuning.

	\textbf{Training Strategy.}
	Training begins with initialization of the auxiliary depth head using $\mathcal{L}_{\mathrm{depth}}$, with the foundation VLM kept frozen.
	The primary VLM is then unfrozen and jointly optimized with the depth head under Eq.~\ref{eq:total_objective}, allowing geometric supervision to refine the visual representations while the frozen twin preserves the foundation model's language distribution.
	After joint adaptation, the auxiliary depth head and twin VLM are discarded.
	For downstream perception and planning, task-specific heads are trained from scratch on the frozen adapted backbone, while language tasks use the adapted VLM directly.
%===============================================================================

\begin{table}[t]
	\centering
	\footnotesize
	\captionsetup{font={footnotesize}, labelfont=bf, labelsep=period, skip=6pt}
	\setlength{\tabcolsep}{4pt}
	\renewcommand{\arraystretch}{1.08}
	\caption{Monocular depth estimation probing on nuScenes.}
	\label{tab:nuscenes_depth}
	\begin{tabular}{@{}lcccccc@{}}
		\toprule
		Metrics & Metric3Dv2~\citep{hu2024metric3d} & UniDepth~\citep{piccinelli2024unidepth} & UniDepthV2~\citep{piccinelli2025unidepthv2} & VGGT~\citep{wang2025vggt} & InternViT & \textbf{InternViT-ours} \\
		\midrule
		$\delta_1$~$\uparrow$ & 84.1 & 84.6 			& \textbf{87.6} & 60.0 & 77.2 & 80.6 \\
		Abs Rel~$\downarrow$  & 23.6 & \textbf{12.7} & 14.3 		    & 28.9  & 19.1 & 16.7 \\
		\bottomrule
	\end{tabular}
	\vspace{-1em}
\end{table}

\begin{table}[t]
	\centering
	\footnotesize
	\captionsetup{font={footnotesize}, labelfont=bf, labelsep=period, skip=6pt}
	\setlength{\tabcolsep}{3pt}
	\renewcommand{\arraystretch}{1.08}
	\caption{3D object detection probing on nuScenes.}
	\label{tab:nuscenes_3d_det}
	\begin{tabular}{@{}lccccccc@{}}
		\toprule
		Metric & PETR~\citep{liu2022petr} & DETR3D~\citep{wang2022detr3d} & FCOS3D~\citep{wang2021fcos3d} & PGD~\citep{wang2022probabilistic} & OmniDrive~\citep{wang2025omnidrive} & InternViT & \textbf{InternViT-ours} \\
		\midrule
		mAP~$\uparrow$ & 37.8 & 34.9 & 34.3 & 36.9 & 35.4 & 34.7 & \textbf{38.1} \\
		\bottomrule
	\end{tabular}
	\vspace{-1em}
\end{table}

\begin{table}[t]
	\centering
	\footnotesize
	\captionsetup{font={footnotesize}, labelfont=bf, labelsep=period, skip=6pt}
	\setlength{\tabcolsep}{2pt}
	\renewcommand{\arraystretch}{1.05}
	\caption{Driving language reasoning evaluation on OmniDrive and nuInstruct. "Specialist Model" denotes domain-specific expert networks trained exclusively on target datasets~\citep{wang2025omnidrive, ding2024holistic}, "Baseline" denotes InternVL3.5-4B optimized via standard SFT. * indicates $\max((Accuracy + mAP + BLEU - MAE)/4, 0)$.}
	\label{tab:Driving VLM Benchmarks}
	\resizebox{\linewidth}{!}{
		\begin{tabular}{@{}llcccccc@{}}
			\toprule
			\multirow{2}{*}{Dataset} & \multirow{2}{*}{Metric} & \multicolumn{2}{c}{Zero-shot models} & \multicolumn{4}{c}{Finetuned models} \\
			\cmidrule(lr){3-4} \cmidrule(lr){5-8}
			& & GPT-4o~\citep{hurst2024gpt} & InternVL3.5-4B~\citep{wang2025internvl3} & VGGDrive~\citep{wang2026vggdrive} & Specialist Model & Baseline & \textbf{Ours-4B} \\
			\midrule
			\multirow{4}{*}{OmniDrive}
			& BLEU~$\uparrow$    & 10.91 & 25.64 & 37.58 & 38.00 & 40.94 & \textbf{42.55} \\
			& ROUGE~$\uparrow$   & 22.34 & 24.90 & 34.40 & 32.60 & 35.98 & \textbf{36.80} \\
			& CIDEr~$\uparrow$   & 24.42 & 24.73 & 86.57 & 68.60 & 86.51 & \textbf{92.09} \\
			& Average~$\uparrow$ & 19.22 & 25.09 & 52.85 & 46.40 & 54.47 & \textbf{57.15} \\
			\midrule
			\multirow{5}{*}{NuInstruct}
			& MAE~$\downarrow$    & 9.93  & 20.56 & 3.08  & 9.08   & 1.67  		    & \textbf{1.44} \\
			& Accuracy~$\uparrow$ & 10.64 & 0.34  & 56.37 & 32.48  & \textbf{59.22} & 58.29          \\
			& mAP~$\uparrow$      & 0.00  & 0.00  & 37.49 & 21.93  & 38.83 		    & \textbf{39.46} \\
			& BLEU~$\uparrow$     & 7.08  & 0.48  & 81.13 & 35.20  & 82.92 		    & \textbf{84.14} \\
			& Average*~$\uparrow$ & 1.95  & 0.00  & 42.98 & 20.13  & 44.83 		    & \textbf{45.11} \\
			\bottomrule
		\end{tabular}
	}
	\vspace{-1.5em}
\end{table}

\begin{table}[t]
	\centering
	\footnotesize
	\captionsetup{font={footnotesize}, labelfont=bf, labelsep=period, skip=6pt}
	\setlength{\tabcolsep}{3pt}
	\renewcommand{\arraystretch}{1.08}
	\caption{Non-reactive trajectory planning evaluation on NAVSIM~\citep{dauner2024navsim}.}
	\begin{tabular}{@{}llcccccccc@{}}
		\toprule
		Method & Base Model & Image & LiDAR & NC~$\uparrow$ & DAC~$\uparrow$ & EP~$\uparrow$ & TTC~$\uparrow$ & Comf.~$\uparrow$ & PDMS~$\uparrow$ \\
		\midrule
		TransFuser~\citep{chitta2022transfuser}       & ResNet-34 	   & \cmark & \cmark & 97.78 		  & 92.63 		   & 78.88  		& 92.89 		 & 99.98 	    & 83.88 \\
		PARA-Drive~\citep{weng2024drive}              & BEVFormer 	   & \cmark & 	     & 97.90 		  & 92.40 		   & 79.30  		& 93.00 		 & 99.80 	    & 84.00 \\
		DRAMA~\citep{yuan2024drama}                   & ResNet-34 	   & \cmark & \cmark & 98.19 		  & 95.18 		   & 81.33  		& 94.17 		 & \textbf{100} & 86.87 \\
		Hydra-MDP~\citep{li2024hydra} 				  & ResNet-34 	   & \cmark & \cmark & 98.30 		  & 96.00 		   & 78.70  		& 94.60 		 & \textbf{100} & 86.50 \\
		DiffusionDrive~\citep{liao2025diffusiondrive} & ResNet-34 	   & \cmark & \cmark & 98.20 		  & 96.20 		   & 82.20  		& 94.70 		 & \textbf{100} & 88.10 \\
		WoTE~\citep{li2025end}                    	  & ResNet-34 	   & \cmark & \cmark & 98.50 		  & 96.80 		   & 81.90  		& 94.90 		 & 99.90 	    & 88.30 \\
		\midrule
		VGGDrive~\citep{wang2026vggdrive}         	  & Qwen2.5-VL-7B  & \cmark &  		 & 98.55 		  & 96.30 		   & 82.92 			& 95.59 		 & 99.98 		& 88.76 \\
		ReCogDrive~\citep{li2025recogdrive}           & InternVL3-8B   & \cmark &  		 & 98.20 	  	  & \textbf{97.80} & 83.50 			& 95.20 		 & 99.80 	 	& \textbf{89.60} \\
		AdaThinkDrive~\citep{luo2025adathinkdrive}    & InternVL3-8B   & \cmark &  		 & 98.50 		  & 94.40 		   & 79.90 			& 94.90 		 & \textbf{100} & 86.20 \\
		\midrule
		Baseline   		 				   		      & InternVL3.5-4B & \cmark &  		 & 97.80 	      & 92.10 		   & 78.30 			& 92.80 		 & \textbf{100} & 83.30 \\
		\textbf{Ours-4B} 				   		      & InternVL3.5-4B & \cmark &  		 & \textbf{98.70} & 96.80 		   & \textbf{83.90} & \textbf{95.80} & \textbf{100} & 89.10 \\
		\bottomrule
	\end{tabular}
	\label{tab:NAVSIM Benchmarks}
	\vspace{-1.5em}
\end{table}

\begin{table}[t]
	\centering
	\footnotesize
	\captionsetup{font={footnotesize}, labelfont=bf, labelsep=period, skip=6pt}
	\setlength{\tabcolsep}{2.5pt}
	\renewcommand{\arraystretch}{1.08}
	\caption{General multimodal capability preservation after driving adaptation.}
	\label{tab:General VLM Benchmarks}
	\begin{tabular}{@{}lccc@{\hspace{8pt}}c@{\hspace{8pt}}ccc@{}}
		\toprule
		\multirow{2}{*}{Dataset} & \multicolumn{3}{c}{InternVL3-8B $\rightarrow$ ReCogDrive~\citep{li2025recogdrive}} & & \multicolumn{3}{c}{InternVL3.5-4B $\rightarrow$ \textbf{Ours-4B}} \\
		\cmidrule(lr){2-4} \cmidrule(lr){6-8}
		& Backbone & Adapted & Diff. & & Backbone & Adapted & Diff. \\
		\midrule
		MMMU~\citep{yue2024mmmu}         & 61.2 & 7.6  & $-$53.6 & & 60.8 & 59.1 & $-$1.7 \\
		MMStar~\citep{chen2024we}        & 60.3 & 21.2 & $-$39.1 & & 60.6 & 61.7 & $+$1.1 \\
		POPE~\citep{li2023evaluating}    & 79.0 & 83.0 & $+$4.0  & & 81.2 & 80.9 & $-$0.3 \\
		OCRBench~\citep{liu2024ocrbench} & 82.9 & 61.5 & $-$21.4 & & 77.6 & 77.4 & $-$0.2 \\
		ScienceQA~\citep{lu2022learn}    & 87.9 & 24.2 & $-$63.7 & & 87.6 & 87.1 & $-$0.5 \\
		\midrule
		\textbf{Avg.}                    & 74.3 & 39.5 & $-$34.8 & & 73.6 & 73.2 & $-$0.3 \\
		\bottomrule
	\end{tabular}
	\vspace{-1.5em}
\end{table}

\begin{table}[t]
	\centering
	\footnotesize
	\captionsetup{font={footnotesize}, labelfont=bf, labelsep=period, skip=6pt}
	\setlength{\tabcolsep}{2.5pt}
	\renewcommand{\arraystretch}{1.18}
	\caption{Ablation of geometric supervision and semantic regularization on InternVL3.5-4B. All variants retain driving SFT; checkmarks indicate active auxiliary objectives. The last row is the full framework.}
	\label{tab:matched_objectives}
	\begin{tabular*}{\linewidth}{@{\extracolsep{\fill}}ccccccccc@{}}
		\toprule
		\multicolumn{2}{c}{Objectives}
		& \multicolumn{3}{c}{Perception}
		& \multicolumn{2}{c}{Driving Language}
		& Planning & General VQA \\
		\cmidrule(lr){1-2}\cmidrule(lr){3-5}\cmidrule(lr){6-7}\cmidrule(lr){8-8}\cmidrule(lr){9-9}
		\multirow{2}{*}{$\mathcal{L}_{\mathrm{depth}}$}
		& \multirow{2}{*}{$\mathcal{L}_{\mathrm{KD}}$}
		& \multicolumn{2}{c}{Depth} & 3D Detection
		& OmniDrive & nuInstruct & NAVSIM & \multirow{2}{*}{Avg.$\uparrow$} \\
		& & $\delta_1\uparrow$ & AbsRel$\downarrow$ & mAP$\uparrow$
		& Avg.$\uparrow$ & Avg.$\uparrow$ & PDMS$\uparrow$ & \\
		\midrule
		$-$ & $-$ & 77.2 & 19.1 & 34.7 & 54.47 & 44.83 & 83.30 & 21.5 \\
		\cmark & $-$ & \textbf{80.8} & \textbf{16.3} & \textbf{38.2} & 56.79 & 44.90 & 86.42 & 21.8 \\
		$-$ & \cmark & 77.3 & 18.9 & 34.9 & 56.27 & 44.83 & 84.56 & 71.0 \\
		\midrule
		\cmark & \cmark & 80.6 & 16.7 & 38.1 & \textbf{57.15} & \textbf{45.11} & \textbf{89.10} & \textbf{73.2} \\
		\bottomrule
	\end{tabular*}
	\vspace{-1.5em}
\end{table}

\begin{table}[t]
	\centering
	\footnotesize
	\captionsetup{font={footnotesize}, labelfont=bf, labelsep=period, skip=6pt}
	\setlength{\tabcolsep}{4pt}
	\renewcommand{\arraystretch}{1.08}
	\caption{Ablation study on the off-policy logit-level KL distillation constraint. $\Delta$ reports gains over w/o KL.}
	\label{tab:ablation_study}
	\begin{tabular}{@{}lcccccc@{}}
		\toprule
		Method & MMMU & MMStar & POPE & OCRBench & ScienceQA & \textbf{Avg.}\\
		\midrule
		InternVL3.5-4B & 60.8 & 60.6 & 81.2 & 77.6 & 87.6 & 73.6 \\
		\midrule
		Ours-4B w/o KL & 7.0  & 4.5  & 67.1 & 21.0 & 9.5  & 21.8 \\
		Ours-4B w/ KL  & 59.1 & 61.7 & 80.9 & 77.4 & 87.1 & 73.2 \\
		\textbf{$\Delta$ w/ KL}
			& \textbf{+52.1} & \textbf{+57.2} & \textbf{+13.8}
			& \textbf{+56.4} & \textbf{+77.6} & \textbf{+51.4} \\
		\bottomrule
	\end{tabular}
	\vspace{-1.5em}
\end{table}

\section{Experimental Results}
\label{sec:result}
\subsection{Implementation Details}
	We build our primary model on InternVL3.5-4B~\citep{wang2025internvl3}. 
	The twin VLM branch is initialized with the same weights and kept frozen throughout training.
	For depth supervision, we attach a lightweight DPT-style depth head~\citep{ranftl2021vision} (30M parameters) to the InternViT visual encoder, and feed multi-scale features from stages 4, 8, 16, and 24 into the depth decoder.
	We optimize the unfrozen parameters of the primary backbone using the AdamW optimizer~\citep{loshchilov2017decoupled} with a base learning rate of $2\times10^{-5}$ for the vision encoder and a cosine decay schedule.
	The overall multi-task loss is balanced via empirical scaling factors: $\alpha=0.25$ for the depth reconstruction objective and $\beta=0.20$ for the off-policy logit-level KL distillation objective, and $\gamma=1.0$ for the driving SFT task loss.
	For the SiLog loss, we set $\lambda=0.85$ and use valid metric depth targets within $[0.5,80]$\,m.

	We construct 170k triplets from the nuScenes training images used by OmniDrive and nuInstruct.
	After DPT head initialization, we jointly train the primary VLM and auxiliary depth head on these triplets and the official driving QA data for two epochs on 16 NVIDIA A100 GPUs (80\,GB each).
	All evaluations share the final jointly adapted VLM checkpoint, with random seed 42 used throughout.

\subsection{Datasets and Metrics}
\label{sec:datasets_metrics}
	Our experiments are structured to evaluate whether our targeted geometric catalyst unlocks a generalized 3D awareness while strictly preserving the model's foundational capabilities. 
	We evaluate across four distinct paradigms spanning perception, language reasoning, non-reactive trajectory planning, and general multimodal understanding. 
	Specifically, camera-only 3D perception is probed on nuScenes~\citep{caesar2020nuscenes} using standard depth and detection (mAP) metrics. 
	Driving-oriented language reasoning is verified on OmniDrive~\citep{wang2025omnidrive} and nuInstruct~\citep{ding2024holistic} via linguistic (BLEU~\citep{papineni2002bleu}, ROUGE~\citep{lin2004rouge}, CIDEr~\citep{vedantam2015cider}) and coordinate-space metrics. 
	Trajectory planning is evaluated on the non-reactive, offline NAVSIM benchmark~\citep{dauner2024navsim} using the Predictive Driver Model Score (PDMS).
	Finally, general open-world capabilities are benchmarked across MMMU, MMStar, POPE, OCRBench, and ScienceQA~\citep{yue2024mmmu, chen2024we, li2023evaluating, liu2024ocrbench, lu2022learn} to monitor catastrophic forgetting.

\subsection{Main Results}	
	\textbf{Spatial Representation Probing on nuScenes.} 
	To verify whether 3D geometric awareness is natively embedded within the visual representations rather than superficially memorized in the text token space, we freeze the optimized visual backbone and evaluate it under strict probing conditions on nuScenes~\citep{caesar2020nuscenes}. 
	We train separate depth and PETR detection heads, following UniDepth~\citep{piccinelli2024unidepth} and the frozen-backbone probing design of DEIMv2~\citep{huang2025real}, respectively.
	As shown in Table~\ref{tab:nuscenes_depth} and Table~\ref{tab:nuscenes_3d_det}, our framework significantly elevates the baseline's latent spatial capacity, driving $\delta_{1}$ depth accuracy from 77.2\% to 80.6\% and 3D detection from 34.7\% to 38.1\% mAP. 
	These results surpass text-based alignment methods (OmniDrive: 35.4\%) and fully fine-tuned domain-specific experts (PETR: 37.8\%), suggesting that implicit depth modeling can produce more transferable 3D features.

	\textbf{Driving-Oriented Language Reasoning.} 
	To evaluate how enhanced spatial awareness translates into multimodal language comprehension, we evaluate our framework on OmniDrive~\citep{wang2025omnidrive} and nuInstruct~\citep{ding2024holistic}.
	These benchmarks challenge models to perform multi-view scene description, spatial topology analysis, and hazard perception.
	As reported in Table~\ref{tab:Driving VLM Benchmarks}, our model consistently sets new state-of-the-art performance across both benchmarks, outperforming the structurally complex competitor VGGDrive~\citep{wang2026vggdrive}.
	This joint success demonstrates that our implicit geometric grounding successfully empowers the language model to construct precise spatial arguments, effectively overcoming the fragile boundary hallucinations typical of text-only coordinate alignment methods.

	\textbf{Non-Reactive Trajectory Planning.}
	To examine whether geometric adaptation also benefits trajectory planning, we train a lightweight planning head on the frozen backbone and evaluate it on NAVSIM~\citep{dauner2024navsim}, following the iPad~\citep{guo2025ipad} experimental setup.
	As revealed in Table~\ref{tab:NAVSIM Benchmarks}, our model achieves a highly competitive PDMS score of 89.10\%. 
	Crucially, despite operating entirely vision-only at inference, our model closely approaches the performance of the twice-as-large 8B parameter model ReCogDrive (89.60\%), while significantly outperforming heavily engineered, LiDAR-reliant expert models such as TransFuser (83.88\%) and WoTE (88.30\%). 
	Furthermore, our model achieves superior safety profiles, including a Non-Collision (NC) score of 98.70\% and a Time-to-Collision (TTC) score of 95.80\%. 
	These results indicate that the learned spatial representations benefit vision-only trajectory planning under non-reactive evaluation.

	\textbf{General Capability Preservation.}
	Standard driving adaptation often triggers catastrophic forgetting due to text-space overfitting on rigid driving terms. 
	We expose this issue by testing models on general VLM benchmarks.
	As shown in Table~\ref{tab:General VLM Benchmarks}, traditional frameworks like ReCogDrive cause a catastrophic degradation in general open-world performance, with its MMMU score plummeting to a near-random 7.6\%, and an average degradation of 34.8 points across benchmarks. 
	In contrast, our model experiences negligible degradation, maintaining competitive scores that closely align with the untuned InternVL3.5-4B baseline. 
	This demonstrates that our tri-stream framework successfully decouples geometric adaptation from semantic preservation, allowing the VLM to remain general-purpose while mastering autonomous driving.

\subsection{Ablation Studies}
	We examine the complementary roles of geometric supervision and semantic regularization by ablating $\mathcal{L}_{\mathrm{depth}}$ and $\mathcal{L}_{\mathrm{KD}}$ under the same training setup, with driving SFT as the baseline (Table~\ref{tab:matched_objectives}).

	\textbf{Effectiveness of Geometric Supervision.}
	Depth supervision improves geometric probing and planning both with and without KL distillation.
	In particular, adding depth to SFT + KL increases 3D detection mAP by 3.2 points and NAVSIM PDMS by 4.54 points.
	These gains support the role of dense depth as a geometric catalyst for downstream perception and planning.

	\textbf{Effectiveness of Semantic Regularization.}
	Depth supervision alone leaves the general-benchmark average at 21.8, whereas adding KL distillation restores it to 73.2 (Table~\ref{tab:ablation_study}).
	The full framework also improves planning over SFT + Depth, despite small reductions in depth and detection performance.
	The two objectives thus play complementary roles: geometric supervision strengthens spatial representations, while the frozen twin preserves the foundation model's open-world capabilities.

%===============================================================================

\section{Conclusion and Limitations}
\label{sec:conclusion}
	In this paper, we introduce Triplet Incremental Training, a novel and non-intrusive paradigm designed to bridge the gap between general vision-language capabilities and spatial awareness in autonomous driving. 
	By leveraging automatically generated (image, depth, text) triplets, our tri-stream architecture enforces explicit geometric constraints via a deployment-discarded transient depth head to ensure zero inference overhead, while simultaneously utilizing a frozen twin VLM as a semantic anchor to prevent catastrophic forgetting.
	Extensive benchmarks across diverse driving tasks confirm that our approach not only achieves state-of-the-art performance but also largely preserves the open-ended reasoning capabilities of the base foundation model.

	\textbf{Limitations.} 
	Despite its competitive performance across multiple benchmarks, a primary limitation of our current validation is its reliance on offline real-world datasets, leaving the framework largely untested in fully interactive, closed-loop simulation environments. 
	In the context of robotic learning, simulation offers an invaluable testing ground characterized by flawless metric depth ground truth, the capacity to synthetically generate vast, diverse, and safety-critical edge cases, and the mechanism for end-to-end closed-loop execution. 
	Such environments are crucial for conducting granular behavioral assessments and isolating specific failure modes under dynamic vehicle-environment interactions. 
	Transitioning our framework into active simulators to evaluate its closed-loop decision-making and robustness represents a vital and immediate avenue for our future research.
%===============================================================================

\clearpage
% The acknowledgments are automatically included only in the final and preprint versions of the paper.
% \acknowledgments{If a paper is accepted, the final camera-ready version will (and probably should) include acknowledgments. All acknowledgments go at the end of the paper, including thanks to reviewers who gave useful comments, to colleagues who contributed to the ideas, and to funding agencies and corporate sponsors that provided financial support.}

%===============================================================================

% no \bibliographystyle is required, since the corl style is automatically used.
\bibliography{submission}  % .bib

\end{document}